\documentclass[10pt,twocolumn,letterpaper]{article}

\usepackage{cvpr}
\usepackage{times}
\usepackage{epsfig}
\usepackage{graphicx}
\usepackage{amsmath}
\usepackage{amssymb}
\usepackage{xspace}
\usepackage{booktabs}
\usepackage{array}
\usepackage{multirow}
\usepackage{subcaption}
\usepackage{enumitem}
\newcolumntype{P}[1]{>{\centering\arraybackslash}p{#1}}

\newcommand{\tea}{\textsc{Teacher}\xspace}
\newcommand{\stud}{\textsc{Student}\xspace}
\newcommand{\full}{\textsc{Teacher-Full}}
\newcommand{\uniform}[1]{\textsc{Teacher-Uniform-}$#1$}
\newcommand{\random}[1]{\textsc{Teacher-Random-}$#1$}
\newcommand{\final}[1]{\textsc{Student-}$#1$\textsc{-Final}}
\newcommand{\inter}[1]{\textsc{Student-}$#1$\textsc{-Intermediate}}

\usepackage[pagebackref=true,breaklinks=true,letterpaper=true,colorlinks,bookmarks=false]{hyperref}

\cvprfinalcopy 

\def\cvprPaperID{0011} 

\ifcvprfinal\fi
\begin{document}

\title{\textit{I Have Seen Enough:} A Teacher Student Network for Video Classification Using Fewer Frames}

\author{\\
Shweta Bhardwaj\\
Indian Institute of Technology Madras\\
{\tt\small cs16s003@cse.iitm.ac.in}
\and
\\
Mitesh M. Khapra\\
Indian Institute of Technology Madras\\
{\tt\small miteshk@cse.iitm.ac.in}
}

\maketitle

\begin{abstract}
Over the past few years, various tasks involving videos such as classification, description, summarization and question answering have received a lot of attention. Current models for these tasks compute an encoding of the video by treating it as a sequence of images and going over every image in the sequence. However, for longer videos this is very time consuming. In this paper, we focus on the task of video classification and aim to reduce the computational time by using the idea of distillation. Specifically, we first train a \textit{teacher} network which looks at all the frames in a video and computes a representation for the video. We then train a \textit{student} network whose objective is to process only a small fraction of the frames in the video and still produce a representation which is very close to the representation computed by the \textit{teacher} network. This smaller student network involving fewer computations can then be employed at inference time for video classification. We experiment with the YouTube-8M dataset and show that the proposed student network can reduce the inference time by upto $30\%$ with a very small drop in the performance. 


\end{abstract}

\section{Introduction}
Today video content has become extremely prevalent on the internet influencing all aspects of our life such as education, entertainment, sports \textit{etc.} This has led to an increasing interest in automatic video processing with the aim of identifying activities \cite{action-recog, video_beyond_short_snippet_classify}, generating textual descriptions \cite{lstm-description}, generating summaries \cite{video_summ}, answering questions \cite{tgif-qa} and so on. Current state of the art models for these tasks are based on the neural encode-attend-decode paradigm \cite{Bahdanau2,Bahdanau}. Specifically, these approaches treat the video as a sequence of images (or frames) and compute a representation of the video by using a Recurrent Neural Network (RNN). The input to the RNN at every time step is an encoding of the corresponding image (frame) at that time step as obtained from a Convolutional Neural Network. Computing such a representation for longer videos can be computationally very expensive as it requires running the RNN for many time steps. Further, for every time step the corresponding frame from the video needs to pass through a convolutional neural network to get its representation. Such computations are still feasible on a GPU but become infeasible on low end devices which have power, memory and computational constraints. 

In this work, we focus on the task of video classification \cite{Youtube8M} and aim to reduce the computational time. We take motivation from the observation that when humans are asked to classify a video or recognize an activity in a video they do not typically need to watch every frame or every second of the video. A human would typically fast forward through the video essentially seeing only a few frames and would still be able to recognize the activity (in most cases). Taking motivation from this we propose a model which can compute a representation of the video by looking at only a few frames of the video. Specifically, we use the idea of distillation wherein we first train a computationally expensive \textit{teacher} network which computes a representation for the video by processing all frames in the video. We then train a relatively inexpensive \textit{student} network whose objective is to process only a few frames of the video and produce a representation which is very similar to the representation computed by the teacher. This is achieved by minimizing the squared error loss between the representations of the student network and the teacher network. At inference time, we then use the student network for classification thereby reducing  the time required for processing the video. We experiment with the YouTube-8M dataset and show that the proposed student network can reduce the inference time by upto $30\%$ and still give a classification performance which is very close to that of the expensive teacher network. 

\begin{figure*}[t]
\centering
\includegraphics[width=12cm, height=5cm]{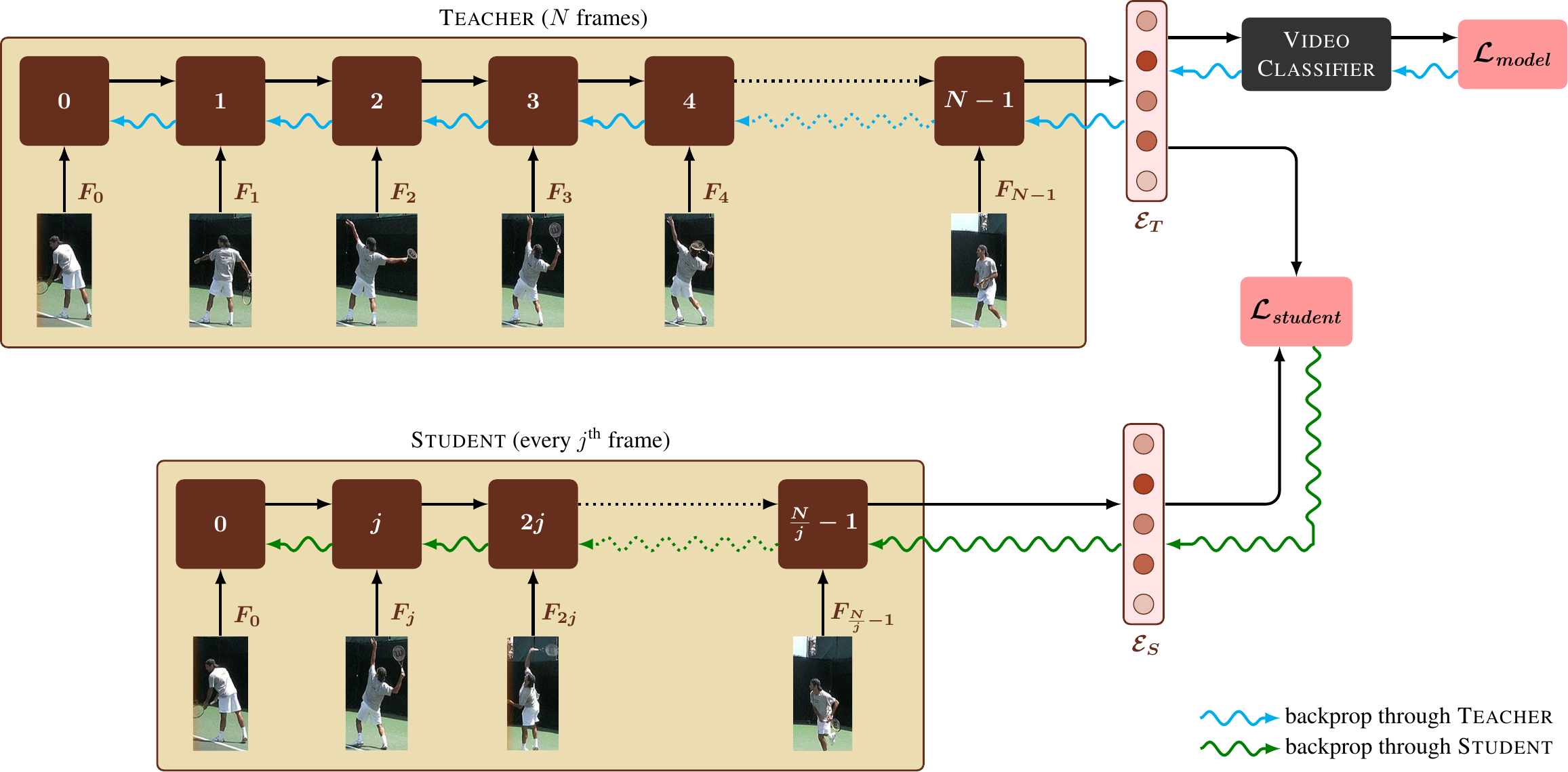}
\caption{\label{diagram} Architecture of \textsc{Teacher-Student} network for video classification}
\end{figure*}

\section{Related Work}
We focus on video classification in the context of the YouTube-8M dataset \cite{Youtube8M}. On average the videos in this dataset have a length of $200$ seconds. Each video is represented using a sequence of frames where every frame corresponds to one second of the video. These one-second frame representations are pre-computed and provided by the authors. The authors also proposed a simple baseline model which treats the entire video as a sequence of these one-second frames and uses an Long short-term memory networks (LSTM) to encode this sequence. Apart from this, they also propose some simple baseline models like Deep Bag of Frames (DBoF) and Logistic Regression \cite{Youtube8M}.
Various other classification models \cite{willow, monkey-typing, temporal-models-yt8m,aggregate-frame-features, deep-models-videos} have been proposed and evaluated on this dataset which explore different methods of: 1) feature aggregation in videos
(temporal as well as spatial) \cite{aggregate-frame-features,willow}, 2) capturing the interactions between labels \cite{monkey-typing} and 3) learning new non-linear units to model the interdependencies among the activations of the network \cite{willow}. 
We focus on one such state of the art model, \textit{viz.}, a hierarchical model whose performance is close to that of 
the best model on this dataset. We take this model as the teacher network and train a comparable student network as explained in the next section.\\
\hspace*{3mm}Our work is inspired by the work on model compression in the context of image classification. For example, \cite{do-deep-really-deep,know-distill,fitnets} use \textit{Knowledge Distillation} to learn a more compact \textit{student} network from a computationally expensive \textit{teacher} network. The key idea is to train a shallow student network to mimic the deeper teacher network, by ensuring that the final output representation and the intermediate hidden representations produced by the student network are very close to those produced by the teacher network. While in their case the teacher and student differ in the number of layers, here, the teacher and student network differ in the number of time steps of frames processed by two networks. 
\section{Proposed Approach}\label{section3}
Our model contains a teacher network and a student network. The teacher network can be any state of the art video classification model but in this work we consider the hierarchical RNN based model. This model assumes that each video contains a sequence of $b$ equal sized blocks. Each of these blocks in turn is a sequence of $m$ frames thereby making the entire video a sequence of sequences. In the case of the YouTube-8M dataset, these frames are one-second shots of the video and each block $b$ is a collection of $m$ such one-second frames. The model contains a lower level RNN to encode each block (sequence of frames) and higher level RNN to encode the video (sequence of blocks). As is the case with all state of the art models for video classification, this teacher network looks at all the $N$ frames of video ($F_{0}, F_{1},\dots,F_{N-1}$) and computes an encoding $\mathcal{E}_{T}$ of the video, which is then fed to a simple feedforward neural network with a multi-class output layer containing a sigmoid neuron for each of the $C$ classes (a video can have multiple labels). The parameters of the teacher network as well as the output layer are learnt using  a standard multi-label classification loss $\mathcal{L}_{model}$, which is a sum of the cross-entropy losses between the true labels $y$ and predictions $\hat{y}$ for each of the $C$ classes, given by:\[ \mathcal{L}_{model} = (- 1) \sum_{i=1}^{C} y_{i} \log(\hat{y}_{i}) + (1- y_{i}) \log(1- \hat{y}_{i}) \]
In addition to this teacher network, we introduce a student network which only processes every $j^{th}$  frame ($F_0, F_j, F_{2j}, \dots, F_{\frac{N}{j}-1} $) of the video and computes a representation $\mathcal{E}_S$ of the video from these $\frac{N}{j}$ frames (which constitutes $\frac{100}{j}$ = $k$ $\%$ of $N$ frames). At the time of evaluation, this representation is fed to the feedforward network with a multi-class output layer. We introduce an additional loss function as shown below which ensures that the representation computed by the student network is very similar to the representation computed by the teacher network.  
\begin{equation*}
 \mathcal{L}_{student} = || \mathcal{E}_{T} - \mathcal{E}_{S} ||^{2}
\end{equation*}
We also try a simple variant of the model, where in addition to ensuring that the final representations $\mathcal{E}_S$ and $\mathcal{E}_T$  are similar, we also ensure that the intermediate representations of the models are similar. In particular, we ensure that the representation of the frames $j$, $2j$ and so on computed by the teacher and  student network are very similar by minimizing the squared error distance between the corresponding intermediate representations. The parameters of the teacher network, student network and output layer are trained jointly as shown in the Figure \ref{diagram}. Note that for ease of illustration, in the figure, we show a simple RNN model as opposed to a hierarchical RNN model. 

\begin{table*}
\centering
 \resizebox{0.58\linewidth}{!}
{
\centering
\begin{tabular}{l|c@{\quad}c@{\quad}c@{\quad}c}
\toprule
\textsc{Model}           & \textsc{AVG-Hit}@1 & \textsc{PERR}  & m\textsc{AP}   & \textsc{GAP} \\
\toprule
\full       & 0.862    & 0.736 & 0.402 & 0.809  \\
\midrule
\uniform{50} & 0.859 &	0.731	& 0.390	& 0.804 \\
\uniform{25}      & 0.855	& 0.725	& 0.371	& 0.798 \\
\uniform{10}         & 0.848    & 0.716 & 0.362 & 0.788 \\
\uniform{5}         & 0.834    & 0.698 & 0.333 & 0.770 \\
\midrule
\random{50}      & 0.841    & 0.702 & 0.305 & 0.775 \\
\random{25}	& 0.832	   & 0.693 & 0.297 & 0.765  \\
\random{10}        & 0.829    & 0.693 & 0.320  & 0.765 \\
\random{5} & 0.804	& 0.665	& 0.288	& 0.731 \\
\midrule
\final{50} & 0.860 &	0.733	& 0.360	& 0.803 \\
\final{25} & 0.857	& 0.727	& 0.385	& 0.802 \\
\final{10} & 0.852    & 0.721 & 0.375 & 0.795 \\
\final{5} & 0.842  & 0.710  & 0.359 & 0.783 \\
\midrule
\inter{50} & 0.862	& 0.739	& 0.385	& 0.805 \\
\inter{25} & 0.851 & 0.718& 0.346&	0.792 \\
\inter{10} & 0.854  & 0.725  & 0.382	& \textbf{0.799}\\
\inter{5} & 0.845 &	0.720	& 0.356	& 0.787 \\
\bottomrule
\end{tabular}
}
\newline
\caption{Performance comparison of proposed \final{k} and \inter{k} models with different baselines on YouTube-8M dataset. Here, \stud-$k$ refers to $k\%$ of frames used by student network. \textsc{Final} encoding and \textsc{Intermediate} encoding refer to two simple variants of the proposed framework.}
\label{table1}

\end{table*}

\section{Experimental Setup}
In this section, we describe the dataset used for our experiments, the hyperparameters that we considered, the baseline models that we compare with and the performance of the two variants of our model.\\

\noindent \textbf{1. Dataset:}
The YouTube-8M dataset \cite{Youtube8M} contains 8
million videos with multiple classes associated with each video. The average length of a video is $200s$ and the maximum length of a video is $300s$. The authors of the dataset have provided pre-extracted audio and visual features for every video such that every second of the video is encoded as a single frame feature. The original dataset consists of 5,786,881 training ($70\%$), 1,652,167 validation ($20\%$) and 825,602 test examples ($10\%$). Since \cite{Youtube8M} does not provide access to the test set, we have reported results on the validation dataset. In this work, we do not use any validation set as we experiment with a fixed set of hyperparameters as explained below.\\

\noindent \textbf{2. Hyperparameters:}
For all our experiments, we used Adam Optimizer with the initial learning rate set to $0.001$  and then decrease it exponentially with $0.95$ decay rate. We used a batch size of $256$. For both the student and teacher networks we used a $2$-layered MultiRNN Cell with cell size of $1024$ for both the layers of the hierarchical model. The size of the hidden representation of the LSTM was 2048. 
For regularization, we used dropout ($0.5$) and $\mathbf{L}_{2}$ regularization penalty of $2$ for all the parameters. We trained all the models for 5 epochs. For the teacher network we chose the value of $m$ (number of frames per block) to be 20 and for the student network we set the value of $m$ to 5. We first train the teacher, student and output layer jointly using the two loss functions described in Section \ref{section3}. After that, we remove the teacher network and finetune the student network and the output layer. \\

\noindent \textbf{3. Evaluation Metrics:} We used the following metrics for evaluating the performance of different models \cite{Youtube8M}:
\begin{itemize}[leftmargin=*,noitemsep]
\item GAP (Global Average Precision): is defined as
\[GAP = \sum_{i=1}^{P} p(i) \nabla r(i)\]
where $p(i)$ is the precision of prediction $i$, $r(i)$ is the
recall of prediction $i$ and $P$ is the number of predictions
(label/confidence pairs). We limit our evaluation to only top-$20$ predictions for each video as mentioned in the YouTube-8M
Kaggle competition.
\item AVG-Hit@$t$ : Fraction of test samples for which the model predicts at least
one of the ground truth labels in the top $t$ predictions. 
\item PERR (Precision at Equal Recall Rate) : For each sample (video), we compute the precision of the top $L$ scoring labels, where $L$ is the number of labels in the ground truth for that sample. The PERR metric is the average of these precision values across all the samples.  
\item mAP (Mean Average Precision) : The mean average precision is computed
as the unweighted mean of all the per-class average precisions.
\end{itemize}
\noindent \textbf{4. Baseline Models:} As mentioned earlier the student network only processes $k\%$ of the frames in the video. We report results with different values of $k$ : $5$, $10$, $25$ or $50$ and compare the performance of the student network with the following versions of the teacher network:
\begin{enumerate}[leftmargin=*,label=\alph*)]
\item \full: The original hierarchical model which processes all the frames of the video.
\item \uniform{k} : A hierarchical model trained from scratch which only processes $k\%$ of the frames of the video. These frames are separated by a constant interval and are thus equally spaced. However, unlike the student model this model does not try to match the representations produced by the full teacher network. 
\item \random{k}:  A hierarchical model trained from scratch which only processes $k\%$ of the frames of the video. These frames are sampled randomly from the video and may not be equally spaced.
\end{enumerate}

We refer to our proposed student network which processes $k\%$ of the frames and matches its final representation to that of the teacher as \final{k}. We refer to the student network which matches all the intermediate representations of the teacher network in addition to the final representation as  \inter{k}.


\section{Results} 
The results of our experiments are summarized in Tables \ref{table1} (performance) and \ref{table2} (computation time). We can show that the observed results are enough to convey the main findings of our work as discussed below. \\
\noindent\textbf{1. Performance comparison against baselines:} As the percentage of frames processed decreases, there is a gap in the performance of \tea and \uniform{50}. However, this gap is not very large. In particular, even when we process only $10\%$ of the frames (\uniform{10}) the drop in AVG-Hit@1, PERR, mAP and GAP is only $2$-$4\%$. As expected, sampling equally spaced frames from the video (\textsc{Uniform}) gives better performance than randomly sampling frames from the video (\textsc{Random}). Further, the gap between the performance of the student network and teacher network is even smaller. In particular, \random{k}$<$ \uniform{k} $<$ \stud-$k$ $<$ \full. This suggests that the student network indeed learns better representations which are comparable to the representations learned by the \tea network. In fact, when we train the student network to match all the intermediate representations produced by the teacher network then we get the best performance. 

\noindent\textbf{2. Computation time of different models:} 
As expected, the computation time of all the models that process only $k \%$ of the frames ($<N$) is much less than the computation time of the teacher network which processes all ($N$) frames of the video (see Table \ref{table2}). We would like to highlight that the \inter{10} gives a drop of $1.2 \%, 0.8 \%, 1.1\%$ and $2 \%$ in GAP, AVG-Hit@1, PERR and mAP scores respectively while the inference time drops by $30\%$.

\vspace*{-4mm}
\begin{table}
\centering
\begin{tabular}{@{}lcccc@{}}
\toprule
\textsc{Model}    & \full & $10\%$ &  $25\% $  & $50\% $     \\
\midrule
\textsc{Time (hrs.)}&  13.00  &  9.11 &  11.00 &   12.50 \\
\bottomrule
\end{tabular}
\vspace*{-3mm}
\caption{Comparison of evaluation time of models using $k\%$ of frames and the \full (original) model  on validation set using Tesla k80 machines}
\label{table2}
\vspace*{-4mm}
\end{table}


\section{Conclusion and Future Work}
We proposed a method to reduce the computation time for video classification using the idea of distillation. Specifically, we first train a teacher network which computes a representation of the video using all the frames in the video. We then train a student network which only processes $k$ \% of the frames of the video. We add a loss function which ensures that the final representation produced by the student is the same as that produced by the teacher. We also propose a simple variant of this idea where the student is trained to also match the intermediate representations produced by the teacher for every $j^{th}$ frame. We evaluate our model on the YouTube-8M dataset and show that the computationally less expensive student network can reduce the computation time by upto $30\%$ while giving similar performance as the teacher network. 

As future work, we would like to evaluate our model on other video processing tasks such as summarization, question answering and captioning. We would also like to experiment with different teacher networks other than the hierarchical RNN considered in this work.


{\small
\bibliographystyle{ieee}
\bibliography{egbib}
}

\end{document}